\documentclass{pbml}

\usepackage{float}
\newcommand{\SIGNIF}{$^\blacktriangle$}

\begin{document}


\title{Linguistically Motivated Vocabulary Reduction \titlelinebreak{} for Neural Machine Translation from Turkish to English}


\institute{label1}{Università degli Studi di Trento, Trento, Italy}
\institute{label2}{Fondazione Bruno Kessler, Trento, Italy}

\author{
  firstname=Duygu,
  surname=Ataman,
  institute={label1,label2},
  corresponding=yes,
  email={ataman@fbk.eu},
  address={Via Sommarive 18, Povo, 38123 Trento, Italy}
}
\author{
  firstname=Matteo,
  surname=Negri,
  institute=label2,
}
\author{
  firstname=Marco,
  surname=Turchi,
  institute=label2,
}
\author{
  firstname=Marcello,
  surname=Federico,
  institute=label2,
}


\shorttitle{Linguistically Motivated Vocabulary Reduction for NMT} 
\shortauthor{D. Ataman et al.} 

\PBMLmaketitle


\begin{abstract}

The necessity of using a fixed-size word vocabulary in order to control the model complexity in state-of-the-art neural machine translation (NMT) systems is an important bottleneck on performance, especially for morphologically rich languages. Conventional methods that aim to overcome this problem by using sub-word or character-level representations solely rely on statistics and disregard the linguistic properties of words, which leads to interruptions in the word structure and causes semantic and syntactic losses. 
In this paper, we propose a new vocabulary reduction method for NMT, which can reduce the vocabulary of a given input corpus at any rate while also considering the morphological properties of the language. Our method is based on unsupervised morphology learning and can be, in principle, used for pre-processing any language pair. We also present an alternative word segmentation method based on supervised morphological analysis, which aids us in measuring the accuracy of our model.
We evaluate our method in Turkish-to-English NMT task where the input language is morphologically rich and agglutinative. We analyze different representation methods in terms of translation accuracy as well as 
the semantic and syntactic properties of the generated output. Our 
method obtains a significant improvement of 2.3 BLEU points over the conventional vocabulary reduction technique, showing that it can provide better accuracy in open vocabulary translation of morphologically rich languages.

\end{abstract}

\section{Introduction}
\label{intro}

Neural machine translation (NMT) is a recent approach to machine translation (MT), which exploits deep learning to directly model the translation probability of texts in two different languages. Although the first models \cite{sutskever2014sequence,bahdanau2014neural} are only few years old, today NMT has already become the new state-of-the-art. Similar to other statistical approaches to MT, NMT is an instance of supervised learning, where a probabilistic model learns to predict an output given the input, based on an history of translation examples. The accuracy of the model is limited by the ability of the system to generalize to unseen examples, which is still an open issue in NMT due to computational restrictions. Current implementations of the model are computationally expensive; they require huge amounts of training time and memory space due to the large number of parameters to optimize. The translation engine uses a word vocabulary whose size is limited in order to control the complexity of the model. However, a text can only be translated if an exact match of the given source word can be found in the vocabulary.

Data sparseness, especially due to rare content words or infrequent inflected word forms, is one of the main reasons that limits the current performance of NMT in low-resourced and morphologically rich languages. For instance, Turkish, the language we focus on in this paper, is an agglutinative language where morphological inflections occur through attachment of suffixes to a given stem. Most syntactic forms in English, such as prepositions, negation, person or copula, are achieved solely through morphological inflections in Turkish. Table \ref{tab:example} illustrates the distance from Turkish to English in terms of the required translations to be generated by an ideal MT system. There are about 30,000 root words and 150 distinct suffixes in Turkish, which can experience agglutinative concatenations and internal changes through fusion to achieve vowel harmony, and cause the morphological tags to grow exponentially \cite{oflazer:2007}.
Hence, the search for alternative word representation techniques that can solve the sparsity problem in Turkish is extremely important and can allow better handling of the input complexity. 

\begin{table}
\begin{small}
 \centering
 \begin{tabular}{ll}
 \hline
 \textbf{Turkish} & \textbf{English} \\
 \hline\hline
 duy(-mak) & \textit{(to) sense} \\
 duygu & \textit{sensation} \\
 duygusal & \textit{sensitive} \\
 duygusalla\c{s}(-mak) & \textit{(to) become sensitive} \\
 duygusalla\c{s}t{\i}r{\i}l(-mak) & \textit{(to) be made sensitive} \\
 duygusalla\c{s}t{\i}r{\i}lm{\i}\c{s} & \textit{the one who has been made sensitive} \\
 duygusalla\c{s}t{\i}r{\i}lamam{\i}\c{s} & \textit{the one who could not have been made sensitive} \\
 duygusalla\c{s}t{\i}r{\i}lamam{\i}\c{s}lardan & \textit{from the ones who could not have been made sensitive} \\ 
 \hline
  \end{tabular}
  \caption{Turkish-to-English translation}
 \end{small}
 \label{tab:example}
\end{table}

Recent studies have tried implicitly extending the vocabulary by segmenting the words in the corpus into smaller units such as characters \cite{ling2015character, lee2016fully}, sub-words \cite{sennrich2015neural, wu2016google} or hybrid \cite{luong2016achieving} units. 
The problem with these approaches is that they disregard any notion of morphology during estimation of the sub-word units, which may lead to loss of semantic and syntactic information preserved in the word structure.
In this paper, we propose to overcome this problem by developing a linguistically motivated segmentation method for open vocabulary translation of morphologically rich languages.  
We present a novel method that can perform segmentation to fit any desired vocabulary size for NMT while also considering the morphological properties of words. 
Being unsupervised, the proposed method can be fundamentally used with any language pair and direction in MT. We evaluate the benefit of our approach in a Turkish-to-English (TR-EN) NMT task against a conventional vocabulary reduction method that relies solely on statistics, and a supervised method that applies segmentation based on morphological analysis. The results show that our linguistically motivated vocabulary reduction method achieves significantly better translation accuracy compared to the conventional method and maintains its performance at different rates of vocabulary reduction. 

\section{Neural Machine Translation}
\label{nmt}

The NMT model we use in this paper is based on the encoder-decoder and attention models described in~\cite{bahdanau2014neural}.
First, a bi-directional RNN (the encoder) maps the sparse one-hot representation of an input sentence $X = (x_1, x_2, \ldots x_m)$ into corresponding dense vectors called encoder hidden states.
Then, a unidirectional RNN (the decoder) step-wisely predicts the target sequence $Y = (y_1, y_2, \ldots y_j \ldots y_l)$ as follows. The $i^{th}$ target word is predicted by sampling from a word distribution computed from the previous target word $y_{i-1}$, the previous hidden state of the decoder, and a convex combination 
of the encoder hidden states (\textit{i.e.} context vector). In particular, each weight of the combination is predicted by the attention model, on the basis of the previous target word, the previous decoder hidden state and the corresponding encoder hidden state. 
Both the encoder and decoder RNNs are implemented with GRU gates~\cite{ChoMBB14}. 
The dimensions of the embeddings and hidden layers are proportional to the vocabulary size. Large vocabularies hence imply more parameters and higher computational costs.

\section{Related Work}
\label{turkishsmt}

In general, two approaches have been proposed to cope with the limited vocabulary problem in NMT. The first one includes purely statistical methods, which aim to predict a set of sub-words that can optimally fit a given vocabulary size. These methods achieved state-of-the-art results for many morphologically rich languages (\textit{e.g.} German, Russian, Czech and Finnish). 

\begin{table}
 \centering
\begin{small}
 \begin{tabular}{lll}
 \hline
 \textbf{Corpus Frequency} & \textbf{Vocabulary Entry} & \textbf{English Translation} \\
 \hline\hline
 1011 & hapishane & \textit{jailhouse} \\
 793 & hapishan@@ & - \\
 587 & hapishanede & \textit{in the jailhouse} \\
 245 & hapishaneden & \textit{from the jailhouse} \\
 229 & hapishanesinde & \textit{at the jailhouse of (him/her/it)} \\
 181 & hapishanenin & \textit{of the jailhouse} \\ 
 100 & hapishanesine & \textit{to the jailhouse of (him/her/it)} \\
 \hline
 \end{tabular}
 \caption{Turkish vocabulary entries obtained with BPE}
 \label{tab:bpe1}
 \end{small}
\end{table}

\begin{table}
\begin{small}
 \centering
 \begin{tabular}{llll}
 \hline
 \textbf{Source} & \textbf{Segmentation} & \textbf{NMT Output} & \textbf{Reference}\\ 
 \hline\hline
 \textit{kanunda} & kan@@ unda & in \textbf{your blood} & in \textbf{the law} \\ \hline
 \textit{sigortalılar} & sigor@@ talı@@ lar & the insure\textbf{r}s & the insure\textbf{d ones}\\
 \hline
 \end{tabular}
 \caption{Translation examples obtained when BPE is applied on Turkish words}
 \label{tab:lemmabpe}
 \end{small}
\end{table}

One such method is Byte-Pair Encoding (BPE), a likelihood-based sub-word unit generation method. BPE is originally a data compression algorithm \citep{gage1994new}, and has been recently modified by \citet{sennrich2015neural} for vocabulary reduction, where the most frequent character sequences are iteratively merged to find the optimal description of the corpus vocabulary. 
Open vocabulary translation using this method is based on the assumption that many types of words can be translated when segmented into smaller units, such as named entities, compound words, and loanwords \cite{sennrich2015neural}.  
Nevertheless, in cases of common morphological paradigms such as the derivational or inflectional transformations which are typically observed in Turkish, the method lacks a linguistic notion which would allow it to better generalize syntactic patterns among the data and use the vocabulary space more effectively. 
Table \ref{tab:bpe1} lists some of the entries found in the NMT dictionary after the segmentation of the corpus with BPE, which stores many repetitions of the same lemma in different surface forms, indicating an inefficacy in capturing a compact representation of the data.  
Another crucial problem is related to the semantic losses which occur due to segmenting words at positions which breaks the morphological structure. Table \ref{tab:lemmabpe} presents some of the typical mistakes observed in the NMT output when BPE is applied for segmentation. In the first example, the Turkish word \textit{kanunda} (translation: \textbf{in the law}), the lemma of which is \textit{kanun} (translation: \textbf{law}), is segmented in the middle of the root, which causes a semantic shift. The segmented word now becomes a completely different word, \textit{kan} (translation: \textbf{blood}). In the second example, segmentation of the suffixes leads to generate the wrong inflected form in English.

Another set of purely statistical methods that attempted to cope with the vocabulary problem in NMT are based on the idea of constructing the translation model directly at the character-level \cite{ling2015character,lee2016fully}. These models use deep neural networks as compositional functions to predict representations of characters and new morphological forms. 
However, these models also assume that, by solely relying on statistics we might be able to capture the morphological rules that form the basics of semantics and syntax of language. Moreover, these models are known to generate spurious words that do not exist in the language \cite{lee2016fully}.

The second family of approaches includes methods that also consider the morphological properties of words but can only reduce the vocabulary to a limited extent, usually by applying cut-off thresholds on the vocabulary and reducing the coverage of the long tail of less frequent words.
For instance, \citet{sanchez2016abu} have used a morphological analyzer to separate words into root and inflection boundaries to achieve vocabulary reduction for NMT. However, in addition to failing to capture a full morphological description of words (\textit{i.e.} generating the complete set of affixes existent in a word), their method cannot reduce the vocabulary of a given text to fit any vocabulary size. Another study tried to overcome this limitation by using the \textit{Baseline} variant of Morfessor \cite{creutz2005unsupervised}, which allows to reach a vocabulary size set prior to segmentation \cite{bradbury2016metamind}. Although providing a sense of morphology into the segmentation process, this tool neglects the morphological varieties between sub-word units, which might result in sub-word units that are semantically ambiguous (\textit{i.e.} either stems or suffixes). 

In conclusion, to our knowledge, there is no vocabulary reduction method for NMT that can both reduce the vocabulary size at any given rate while also considering the individual morphological properties of the generated sub-word units. We aim to solve this problem with the segmentation method described in the next section. 

\section{Linguistically Motivated Vocabulary Reduction}

We present a linguistically motivated segmentation method that achieves open vocabulary translation while considering the morphological properties of individual sub-word units. First, we propose using a supervised segmentation method based on morphological analysis, which helps us to evaluate our vocabulary reduction technique in terms of its ability to generalize the morphology of language from input data. This method aims to represent words in a less sparse way while preserving the complete morphological information. Later, we describe the method proposed in this paper, an unsupervised morphology learning algorithm that predicts the sub-word units in a corpus by a prior morphology model while reducing the vocabulary size to fit a given constraint. 

\subsection{Supervised Morphological Segmentation}
\label{analyzer}

As a supervised approach to linguistically motivated segmentation, we use a me\-thod which can reduce the word vocabulary of the Turkish corpus to only the root words along with a set of suffix units that are represented in terms of their inflectional roles. This representation maintains a full description of the morphological properties of sub-word units in a word while minimizing the sparseness caused by inflection and allomorphy.
We adopt the pre-processing approach of \citet{bisazza2009morphological}, who used the suffix combinatory finite-state analyzer of \citet{oflazer1994two} to tag each sub-word unit in a  Turkish word, and a morphological disambiguation tool \cite{sak2007morphological} to decrease the sparseness caused by suffix allomorphy.  After the pre-processing, we separate all roots and suffix tags into separate tokens and add an end-of-word (EOW) symbol for each analyzed word. 

\subsection{Unsupervised morphological segmentation}
\label{method}

Supervised methods can provide the best accuracy in analysis, although, an ideal approach for MT should not require language-specific resources. 
Therefore, in this paper, we suggest to extend the unsupervised morphology induction framework Morfessor to develop a novel linguistically motivated vocabulary reduction method in NMT, which optimizes the complexity of the segmentation model with a constraint on the vocabulary size. The analysis of \citet{creutz2005inducing} shows that Morfessor models optimized with the Maximum A-Posteriori (MAP) criterion generally achieve the best results. Our model is based on Morfessor \textit{Flatcat} \cite{gronroos2014morfessor}, a variant of this model family that uses a category-based Hidden Markov Model (HMM) and a flat lexicon structure. 
The category-based model is essential for a linguistically motivated segmentation as words would only be split considering the possible categories of their sub-words, preventing to split the words at random positions when a frequent sub-word is observed. 

The aim of MAP optimization is to avoid overfitting by finding a balance between model accuracy and complexity.
The model consists of two parts, a morpheme lexicon and a grammar that combines the language units together and generates new words. The MAP estimate of the overall system is given as:
\begin{equation}
    \label{eq:MAP}
    M^* = argmax_M  P(D|M)  P(M) 
\end{equation}
\noindent
where the two factors represent the likelihood of the training corpus \textit{D} and 
the prior probability of the model \textit{M}. The former is estimated by an HMM which considers transitions between different morpheme categories (\textit{e.g.} stem to suffix) when a word is constructed. The latter is modeled considering individual properties of the generated morphemes $\mu_i$:
\begin{equation}
    P(M) \approx m! \prod_i^m P(usage(\mu_i))P(form(\mu_i))
\end{equation}
where \textit{m} is the number of distinct morphemes in the lexicon \cite{creutz:2007}. The \textit{usage} of a morpheme is related to its meaning and is modeled with its frequency, length, and the left and rightward perplexities. The \textit{form} of a morpheme is the set of physical properties that distinguish it from the others in the lexicon. 

Using the a-posteriori probability, one can train a segmentation model considering both the model complexity and the maximum-likelihood of the corpus, without any control on the size of the output lexicon. 
In order to use the model to achieve controlled vocabulary reduction for NMT, 
we insert a constraint on the desired lexicon size into the MAP optimization 
by applying a regularization weight over the lexicon cost and giving more favor in a reduction of the model complexity during optimization. The cost function is then estimated by the general formula:

\begin{equation}
 L(D,M) = - log P(D|M) - \alpha log P(M) 
 \end{equation}

where a higher $\alpha$ would force the algorithm to generate a smaller lexicon size and a higher amount of segmentation. Considering the tendency of the flat lexicon models to keep the frequent words unsegmented in the corpus \cite{gronroos2014morfessor}, in order to achieve a more accurate segmentation model we disregard the frequency distribution P($\mu_i$) from the weighted part of the cost function. In fact, the value of the term is generally too small to affect the model complexity, but has an important role in determining the characteristics of the discovered morphemes. 

For a given NMT vocabulary size limit, by setting the regularization weight $\alpha$ as $\frac{m_1}{m_2}$, where \textit{$m_1$} is the initial vocabulary size of the corpus, and \textit{$m_2$} is 
the desired vocabulary size, we achieve the right amount of regularization and the output lexicon size. The modified model has a new input parameter, \textit{output lexicon size}, which sets the amount of regularization that reduces the vocabulary to the desired size. By using the parameter as a convergence limit we also minimize the model convergence time. 

\section{Experimental Set-up}
\label{experiments}

We design two sets of experiments in order to evaluate our method. In the first experiment, we evaluate its ability to capture the morphological properties of sub-word units. As an indicator of vocabulary reduction that maintains the full morphological description and semantics of the original word, we deploy the supervised segmentation described in Section \ref{analyzer}. However, the supervised method can only reduce the vocabulary to an extent. Hence, to eliminate the effect of out-of-vocabulary (OOV) words in test set to the accuracy, we set-up a controlled environment where we segment the data using the supervised method and sample the training, development and test sets so that they do not contain any OOVs. We also compare the performance of the method presented in Section \ref{method}, and BPE-based segmentation on the same data sets, and the case without segmentation. 
In order to achieve a fair comparison between the two vocabulary reduction methods, we train the splitting rules of our method and BPE only on the source side of the parallel data. In the second set of experiments, we evaluate our method in a real case scenario. We do not include the supervised method in this phase as its performance would be highly affected by the amount of OOVs in the training and test sets. In Experiment 2.a, we use data sets of similar distribution, whereas in Experiment 2.b, we increase data sparsity by adding generic data to the training set. We segment the source side of parallel corpora using different methods while we segment the target side with BPE. We measure the performance in either experiment (2.a and 2.b) on the same test set. 

We use two sets of data for training our NMT systems. The first data set is the Turkish-English portion of TED Talks \cite{cettoloEtAl:EAMT2012} from IWSLT \cite{iwslt10:EC:overview} and is used in Experiment 1 and 2.a. The second data set is a combination of TED Talks and a collection of generic data from EU Bookshop \cite{SkadinsEA:LREC14}, Global Voices, Gnome, Tatoeba, Ubuntu \cite{TIEDEMANN12.463}, KDE4 \cite{tiedemann2009news}, Open Subtitles \cite{lison2016opensubtitles2016} and SETIMES \cite{tyers2010south}, filtered using the invitation model of \citet{cuong2014latent} to reduce the size. The generic data is used in Experiment 2.b. In all the experiments, we use development and test sets of 1,000 sentences and use the remaining data for training the models. The statistics of all the data sets used in each experiment are given in Table \ref{tab:data}. 

\begin{table}[t]
\begin{small}
 \centering
 \begin{tabular}{ccccc}
 \hline
 \textbf{Data set} & \textbf{Experiment} &\textbf{\#sentences (K)} & \textbf{\#tokens (M)}& \textbf{\#types (K)}\\
 \hline\hline
 TED &(1) & 115 & 1.6 (TR) - 2.2 (EN) & 141 (TR) - 44 (EN) \\
 TED & (2.a) & 133 & 1.9 (TR) - 2.7 (EN) & 169 (TR) - 53 (EN)\\
 TED + Generic & (2.b) & 283 & 4.1 (TR) - 5.6 (EN) & 268 (TR) - 96K (EN) \\
\hline
 \end{tabular}
\quad
\begin{tabular}{|c|cccc|}
 \hline
 \end{tabular}
 \caption{Data sets used in each experiment. K - thousand, M - million.}
 \label{tab:data}
 \end{small}
\end{table}

The NMT models used in the evaluation are based on the Nematus toolkit \cite{nematus}. They have a hidden layer and embedding dimension of 1024, a mini-batch size of 100 and a learning rate of 0.01. The dictionary size is 40,000 (\textit{src} \& \textit{trg}) in the $1^{st}$, and 30,000 (\textit{src}) - 40,000 (\textit{trg}) in the $2^{nd}$ experiment. We train the models using the Adagrad \cite{duchi2011adaptive} optimizer with a dropout rate of 0.1 (\textit{src} \& \textit{trg}) and 0.2 (\textit{embeddings and hidden layers}). We shuffle the data at each epoch. BPE merge rules are of equal size to the dictionary.  
We train the models for 50 epochs and choose the best model on the development set for translating the test set. 

The modified Morfessor \textit{FlatCat} models \cite{gronroos2014morfessor} are trained with a perplexity threshold of 10, a length threshold of 5, and an \textit{output lexicon size} of 40,000 (\textit{Experiment 1 \& 2.a}) and 30,000 (\textit{Experiment 2.b}), which is a new input parameter added to the model implementation. Training time is 20 minutes (using an Intel Xeon E3-1240 v5 CPU), while segmentation time varies from 10 to 30 minutes, depending on the corpus size. 
Performance is measured using the BLEU~\cite{bleu}, TER~\cite{ter} and CHRF3~\cite{chrf} scores and significance tests are computed with Multeval~\cite{multeval}.

\section{Results and Discussion}

Table \ref{result:ted} shows the performance of different segmentation methods in Experiment 1. Our linguistically motivated vocabulary reduction (LMVR) method achieves the best performance on average, proving our hypothesis that a correct morphological representation generates more accurate translations. Our method outperforms the strong baseline of BPE-based segmentation by \textbf{2.2} BLEU, \textbf{4.8} TER and \textbf{1.6} CHR3F points. 
The performance is slightly higher than the supervised method, which is related to the ambiguity caused by loss of information during the morphological analysis. 
The predicted vocabularies also indicate the significant difference between LVMR and BPE, where 73\% of the sub-word units in the vocabulary are completely different. In order to better illustrate the properties of the generated sub-word units, we present example translations of two words from the test set. The two words have different roots, the first one is \textit{ağ} (translation: \textbf{net}), and the second one is \textit{ağla} (translation: \textbf{(to) cry}). BPE segments both words to the same root \textit{ağ}, a character sequence frequently observed in root words in Turkish.  
In the first case, both unsupervised methods segment the word into the same sub-word units, while the embedding of the sub-word unit segmented with BPE is semantically ambiguous and generates unreliable translations. On the other hand, our method can preserve the correct meaning in both cases. 

\begin{table}[t]
\begin{small}
\begin{centering}
\begin{tabular}{cccc}
 \hline
 \multicolumn{4}{c}{\textbf{1. TED corpus, no-OOV case, voc=40K}}\\\hline\hline
 \textbf{Method} & \textbf{BLEU$\uparrow$} & \textbf{TER$\downarrow$} & \textbf{CHRF3$\uparrow$}\\
 \hline
 No Segmentation & 17.77 & 68.07 & 38.94 \\
 \hline
  BPE & 19.52 & 66.23 & 42.33\\
  \hline
  Supervised & 21.61\SIGNIF & 61.76\SIGNIF & \textbf{44.01}\\ 
  \hline
  \textbf{LMVR} & \textbf{21.71}\SIGNIF & \textbf{61.41}\SIGNIF & 43.90 \\
  \hline
 \end{tabular} 
\quad
\begin{tabular}{cccc}
 \hline
 \textbf{Input (\textit{Reference})} & \textbf{Method} &\textbf{Segmentation} &  \textbf{Output} \\
 \hline\hline
 ağlarını  & BPE & ağ@@ larını & the cry \\
 (\textit{the nets}) & LMVR & ağ +larını & the nets  \\
 & Supervised & ağ +Noun + A3pl <EOW> & networks \\
 \hline
 ağlamayacak   & BPE & ağ@@ lamayacak & will not survive \\
 (\textit{would not be crying}) &  LMVR & ağlama +yacak & will not cry  \\ 
 & Supervised & ağla +Neg +Fut +A3sg <EOW>& will not cry  \\
  \hline
 \end{tabular} 
 \caption{Results of Experiment 1 - TED corpus and no-OOV case. Top: Output accuracies, where \SIGNIF indicates statistically significant improvement over the BPE baseline (p-value~$<$~0.05). Bottom: Translation examples.} 
 \label{result:ted}
\end{centering}
\end{small}
\end{table}

\begin{table}[h]
\begin{small}
\begin{centering}
 \begin{tabular}{ccccccc}
 \hline
 & \multicolumn{3}{c}{\textbf{2.a TED corpus, OOV case, voc=40K}} & \multicolumn{3}{c}{\textbf{2.b Large corpus, OOV case, voc=30K}}\\
 \hline\hline
  \textbf{Method} & \textbf{BLEU$\uparrow$} & \textbf{TER$\downarrow$} & \textbf{CHRF3$\uparrow$} & \textbf{BLEU$\uparrow$} & \textbf{TER$\downarrow$} & \textbf{CHRF3$\uparrow$}\\
 \hline
  BPE & 20.45 & 64.50 & 42.65 & 24.42 & 60.14 & 47.05 \\
 \hline
  \textbf{LMVR}  & \textbf{22.76}\SIGNIF & \textbf{62.94}\SIGNIF & \textbf{45.36} & \textbf{25.42}\SIGNIF & \textbf{58.88}\SIGNIF & \textbf{47.71} \\
  \hline
 \end{tabular} 
 \caption{Results of Experiment 2 - OOV presence and different rates of vocabulary reduction. \SIGNIF indicates statistically significant improvement over the BPE baseline (p-value~$<$~0.05).} 
 \label{result:2}
 \end{centering}
\end{small}
\end{table}

In Experiment 2, we evaluate our method at different rates of vocabulary reduction according to the vocabulary sizes given in Table \ref{tab:data}. All metrics confirm that our method achieves better performance than the baseline in both experiments. In Experiment 2.a, at a vocabulary reduction rate of 4.25 (170K -> 40K), we obtain an improvement of \textbf{2.3} BLEU points over the baseline. In the most challenging case, Experiment 2.b, we increase the training set using data coming from varying domains, which maximizes the sparseness due to rare word forms in the corpus. Furthermore, we decrease the source vocabulary limit to 30,000, requiring a vocabulary reduction rate of 9 (270K -> 30K). As given in Table \ref{result:2}, our method can still outperform the baseline by \textbf{1.0} BLEU point. The results and the computational efficiency of our method prove that it can be deployed in practical NMT systems trained with generic corpora. 

\section{Conclusion}

In this paper we have addressed the vocabulary limitation in NMT, which has been an open issue in the translation of morphologically rich languages. For this purpose, we have proposed a novel linguistically motivated vocabulary reduction method that can achieve open vocabulary translation while, unlike previous approaches, maintaining a linguistic notion at the sub-word level. The method is completely unsupervised and can estimate a fixed size dictionary of sub-word units considering their individual morphological properties. We have evaluated our method against a statistical vocabulary reduction method and showed that our method obtains significantly better performance due to bringing a linguistic notion into the segmentation process.

\section*{Acknowledgements}
This work has been partially supported by the EC-funded H2020 projects QT21 (grant no.  645452) and ModernMT (grant no. 645487).
The authors would like to thank Arianna Bisazza and Prashant Mathur for their contributions to this study.

\bibliography{mybib}


\correspondingaddress
\end{document}